# Distributed Swarm Collision Avoidance
# Based on Angular Calculations

SeyedZahir Qazavi [1], Samaneh Hosseini Semnani [2]

*Abstract*— **Collision avoidance is one of the most important topics in the robotics field. The goal is to move the robots from initial locations to target locations such that they follow shortest non-colliding paths in the shortest time and with the least amount of energy. In this paper, a distributed and real-time algorithm for dense and complex 2D and 3D environments is proposed. This algorithm uses angular calculations to select the optimal direction for the movement of each robot and it has been shown that these separate calculations lead to a form of cooperative behavior among agents. We evaluated the proposed approach on various simulation and experimental scenarios and compared the results with FMP and ORCA, two important algorithms in this field. The results show that the proposed approach is at least 25% faster than ORCA and at least 7% faster than FMP and also more reliable than both methods. The proposed method is shown to enable fully autonomous navigation of a swarm of crazyflies.**

## I. INTRODUCTION

In recent years, multi-robot systems have been widely welcomed by the research communities thanks to their widespread and growing use in various fields such as, moving objects in large warehouses using robots [1], group movement of agents in animations or computer games [2], intelligent and autonomous public transportation, ship navigation in large ports with heavy traffic [3], drug delivery in blood vessels by nanobots [4], aerial shows and light painting at night [5], etc. These are just a few sample applications of multi-robot systems.

One of the main challenges in some multi-robot applications is finding an optimal collision avoidance/motion planning algorithm that can move properly each agent along with others. Motion planning is the problem of finding an appropriate sequence of commands to move an agent from one point to another. Correspondingly, swarm motion planning is the problem of finding multiple such sequences in a multi-agent system as illustrated in Fig. 1. Collective motion planning can be divided into two parts. First of all, finding the optimal assignment between agents' initial and final locations in order to minimize the possibility of collision in straight paths from initial to final positions. Secondly, solving collisions that may occur between two or more agents when they are moving toward their pre-assigned goal locations. In this paper, we focus on the second part. For the first part, different assignment algorithms can be used, e.g. Hungarian [6] algorithm can find optimal assignment to minimize the probability of collision.

On the other hand, motion planning algorithms can be divided into two main categories: centralized and distributed approaches. In centralized methods, all movement calculations are performed on a central system and it is assumed that this system has all agents' information and issues position and velocity commands at each time step for all agents. Most centralized algorithms look at the motion planning problem as an optimization problem. They first introduce a cost function and then try to optimize it using various optimization techniques. For example [7] uses Particle Swarm Optimization (PSO), [8] uses Mixed Integer/Linear Programming (MILP) and [9] uses Sequential Convex Programming (SCP).

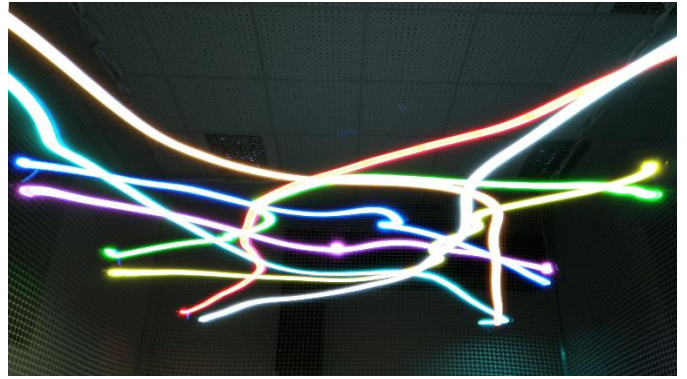

**Fig. 1** Light painting of 8 crazyflie 2.0 trajectories. In this test, 8 crazyflies are located in two parallel lines. Each crazyflie is moving toward its initial position symmetrical point relative to the center.

In general, centralized methods provide more efficient solutions due to their comprehensive knowledge of the environment and agents, but they have limitations such as computational load, simultaneous communication, and sensitivity to environmental changes that make their implementation in real applications difficult or impossible.

The main objective of distributed methods, is to delegate the calculations of the motion of each agent to itself. As a result, this group hopes to overcome most of the limitations of the centralized methods.

Velocity Obstacle (VO) is a representative name for a large group of distributed motion planning methods [10-18] that their original idea was introduced in [10]. In this method, each robot creates a cone according to the position of each of its neighbors and tries to select its velocity vector outside these cones. One of the main problems of the basic VO [10] is decision making in reciprocal situations. When two agents facing each other, they make oscillating and non-optimal movements. To solve this issue, Reciprocal Velocity Obstacle (RVO) is introduced [17]. Finally, Optimal Reciprocal Collision Avoidance (ORCA), an improved version of RVO, presented in [18] to eliminate oscillations of the obtained paths by changing the shape of the avoidance conical space and adding time horizon. ORCA is a well-known, real-time and optimal swarm collision avoidance algorithm for two-dimensional spaces. All VO family methods require many information about the neighbors. Each agent needs to know its neighbors' velocity vector at each time step to determine its movement's direction for the next step. This can be an important

[1] SeyedZahir Qazavi is with the Department of Electrical and Computer Engineering, Isfahan University of Technology, Isfahan 84156-83111, Iran (e-mail: sz.qazavi@ec.iut.ac.ir)
[2] Samaneh Hosseini Semnani is with the Department of Electrical and Computer Engineering, Isfahan University of Technology, Isfahan 84156-83111, Iran (e-mail: samaneh.hosseini@iut.ac.ir)

implementation constraint in many applications.

Another large family of motion planning methods is Potential Field (PF). In this category which was initially designed to single robot path planning [19], the robots have a repulsive force relative to each other and to obstacles, and each robot has a gravitational force to its target coordinates. The resultant and interaction of these forces over the time leads to the convergence of the system. Force-based Motion Planning (FMP) [5] can be considered as one of the newest and most powerful methods of the PF family. In FMP, repulsive and navigational forces are calculated based on the equations calculated in the semi-flocking algorithm [20]. FMP is real-time, low-cost and faster than ORCA, but in complex scenarios, its convergence speed is much reduced.

Several distributed methods have also been proposed based on reinforcement learning [21-24]. These methods are simpler to design and implement and are more scalable than classical and heuristic methods, but their main drawback is uncertainty. In these methods, there is no guarantee of non-collision and there will be a possibility of error and collision in the system at any time.

This paper presents Angular Swarm Collision Avoidance (ASCA), a new algorithm for motion planning of large agent teams. ASCA is distributed, real-time, low cost and based on holonomic robots in both two- and three-dimensional space. In this algorithm, each agent calculates its movement's direction based on its own sense (knowing relative position of other agents/obstacles) at each time step i.e. each agent does not need to know the state of neighboring agents. The proposed method calculates a possible interval for its movement in each step and then quantifies its velocity size and direction based on that. ASCA is parameter-free and only needs robot and environment constraints e.g. maximum allowed speed of each agent and minimum possible separation distance between the agents. It is shown that ASCA is faster in simulation compared to state-of-the-art algorithms ORCA and FMP.

ASCA is similar to ORCA in which they both are collision avoidance algorithms. In both algorithms agents move toward their goal locations until facing a collision. In contrast, ASCA and ORCA are different in terms of how they perceive the environment and how they make decisions. In this way, ASCA requires much less knowledge of its surroundings and agents make decisions using location, but in ORCA agents use the velocity vectors to plan their next move. Compared to FMP, the method introduced in this article works quite differently. Both methods are designed based on the least possible information from the environment, but how they work is completely different. ASCA decision making is direct and each agent calculates its movement direction according to its environment sensing, but FMP acts indirectly and uses the relationships of repulsive and gravitational forces to calculate the total force to Move in that direction. ASCA against FMP can operate optimally without having to specify additional parameters based on problem.

The rest of this paper is organized as follows. Section II presents ASCA, the proposed motion planning algorithm based on avoiding collisions using angular calculations. Section III compares ASCA simulation results against other state of the art methods. Section IV presents experimental results of ASCA in 3D space. Finally, section V concludes the paper and presents some future directions.

## II. Approach

The main idea of most collision avoidance methods is the same: moving straight towards the target until facing a collision. ASCA deals with the problem in the same way but the differences are in the time, place and decision-making algorithm that it starts to react to avoid collision. In this paper, all agents use the same method to make decisions and at the end, the collective movement of the system is shaped in such a way that each of them moves properly in cooperation with its neighbors.

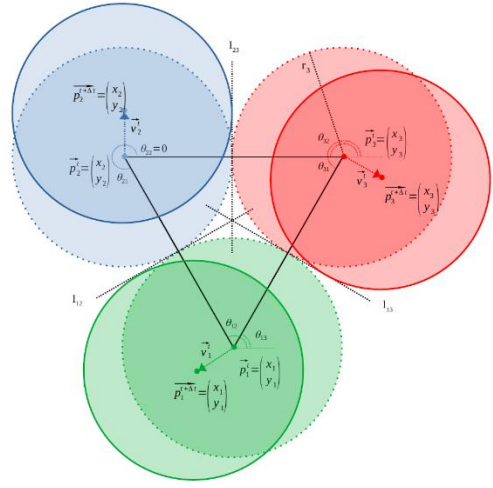

**Fig. 2** Angular calculations in ASCA on the presence of 3 agents $a_1$, $a_2$ and $a_3$ at time t with $r_1$, $r_2$ and $r_3$ avoiding space.

In ASCA, all robots move toward their target independently of others, and if they reach another robot or obstacle, each robot defines a range of motion for itself and continues to move accordingly.

### A. Problem Statement

Motion planning problem deals with a group of agents moving from predefined initial positions to assigned final positions avoiding robot-robot and obstacle-robot collisions.

### B. Assumption

| | |
|---|---|
| N | Number of agents |
| $a_i$ | Agent $i$ |
| $\overrightarrow{I_i}$ | Initial coordinates of $a_i$ |
| $\overrightarrow{F_i}$ | Destination coordinates of $a_i$ |
| $\overrightarrow{p_i^t}$ | Location coordinates of $a_i$ at time t |
| $\overrightarrow{v_i^t}$ | Velocity vector of $a_i$ at time $t$ |
| $\Delta t$ | Length of time steps throughout the algorithm execution |
| $t_{final}$ | Maximum allowed travel time |
| $d_{max}^t$ | Maximum distance to goal position among all agents until time $t$ |
| $d_{final}$ | Maximum distance that agents can have from their target position to terminate the algorithm |
| $r_i$ | Radius of avoiding circular (or spherical) space around $a_i$ |
| $d_{min}^{ij}$ | Minimum required separation distance between $a_i$ and $a_j$ |
| $v_{max}^i$ | Maximum allowed velocity for $a_i$ |

Maximum velocity and collision radius can be selected separately for each agent. For simplicity, in the rest of this paper it is assumed that all agents have the same structure, so we have:

$$\forall i: v_{max}^i = v_{max} \text{ and } r_i = r$$
$$\forall i,j: d_{min}^{ij} = d_{min} = 2r \tag{1}$$

ASCA can be implemented in both two and three-dimensional environments:

### C. 2-D Environment

Fig. 2 represents a collision situation by tree agents. None of the agents should enter their neighbor's avoiding circle e.g. $a_1$ cannot cross the two lines $l_{12}$ and $l_{13}$. To avoid collision with $a_2$, $a_1$ needs to select its velocity vector direction from the feasible angle range $R_{12}$ defined in (4). Similarly, $R_{13}$ can be calculated as the feasible angle between $a_1$ and $a_3$.

$$l_{12} \perp \overrightarrow{p_1 p_2} \tag{2}$$

$$\theta_{12} = \tan^{-1}\left(\frac{y_2 - y_1}{x_2 - x_1}\right) \tag{3}$$

$$R_{12} = \left[\theta_{12} + \frac{\pi}{2}, \theta_{12} + \frac{3\pi}{2}\right] \tag{4}$$

The intersection of $R_{12}$, $R_{13}$ determines the total feasible motion range for $a_1$. This can be extended for $a_i$ with $m$ neighbors at time step t, as:

$$R_i^t = \bigcap_{j=1}^{m} R_{ij}^t \tag{5}$$

The angle in the trigonometric circle is periodic and the beginning value of the feasible angle interval can be larger than the end of that. So, in this case, we change the angle interval $[\alpha_1, \alpha_2]$ to $[\alpha_1, \alpha_2 + 2\pi]$.

Equation (6) shows intersection calculation for this case:

$$\begin{aligned} R &= [\alpha_1, \alpha_2] \cap [\beta_1, \beta_2] \\ &= [\max(\alpha_1, \beta_1), \min(\alpha_2, \beta_2)] \end{aligned} \tag{6}$$

Where $\alpha_1, \alpha_2, \beta_1, \beta_2$ are four hypothetical radian angles.

After calculating feasible range, each agent selects an appropriate velocity vector direction in this range. If the interval is empty, it means that any direction leads to collision, so the velocity vector has to be set to zero.

The size of velocity vector is selected based on the robot structure ($v_{max}$), problem definition space and distance from the target position. To determine the velocity direction, according to Equations (7) to (9), after calculating feasible angle range, if the direct line of sight to the target is in this interval, then the agent moves directly towards its goal position, otherwise, it has to select the closest possible direction to the target which is also inside its calculated interval i.e. the agent has to move tangentially into one of the two boundaries of the range. To have similar behavior among all agents, always the lower limit of the feasible range is selected.

$$\overrightarrow{d_i^t} = \overrightarrow{F_i} - \overrightarrow{p_i^t} \tag{7}$$

$$\theta_i^t = \begin{cases} \angle\overrightarrow{d_i^t}, & \angle\overrightarrow{d_i^t} \in R_i^t \\ \min(R_i^t), & o.w. \end{cases} \tag{8}$$

$$\overrightarrow{v_i^t} = \left|\overrightarrow{d_i^t}\right| \angle \theta_i^t \tag{9}$$

Where $d_i^t$ is $a_i$ line of sight vector toward its destination position at time $t$, $\angle d_i^t$ is angle of vector $d_i^t$, $R_i^t$ is no-collision angle interval for $a_i$ at time $t$ and $\theta_i^t$ is final chosen movement direction for $a_i$.

If $|v_i^t| > v_{max}$ i.e. $a_i$ violates the $v_{max}$ constraint, then we limit the velocity applying Equation (10).

$$\overrightarrow{v_i^t} = v_{max} \frac{\overrightarrow{v_i^t}}{|\overrightarrow{v_i^t}|} \tag{10}$$

### D. 3-D Environment

The only difference between 2D and 3D versions of ASCA is that in the two-dimensional version, the avoidance space around each agent is a circle, the boundaries are lines and the vectors are two-dimensional. But in the three-dimensional environments, the avoidance space is spherical, the boundaries are planes and the vectors have three-dimensions. As shown in Fig.3, the projection of a sphere on each coordinate planes XY, XZ and YZ is a circle. Each agent based on its neighbors, calculates three feasible angle range each on one of the projected planes. Then three velocity vectors $(v_i^t)^{xy}$, $(v_i^t)^{xz}$ and $(v_i^t)^{yz}$ will be calculated each on one of the coordinate planes. As represented in Equation (11), each two-dimensional vector consists of two independent orthogonal components. Each three-dimensional vector consists of three independent orthogonal components, Equation (12) shows how by combining the components of the three 2D velocity vectors, the final 3D velocity vector is calculated.

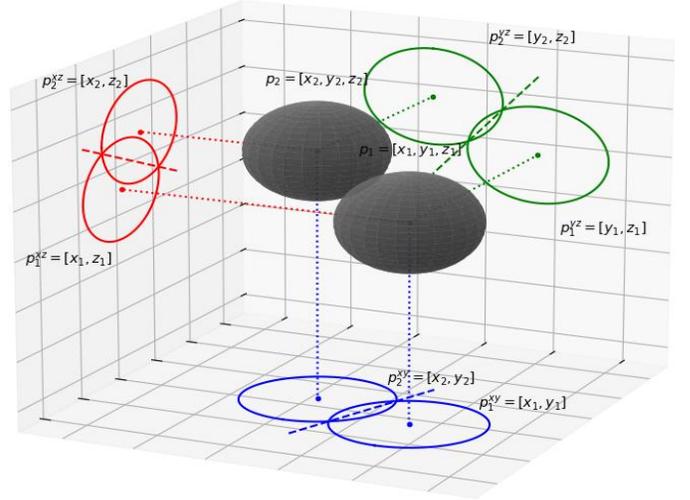

**Fig. 3** There are 2 agents at $\boldsymbol{p_1}$ and $\boldsymbol{p_2}$ coordinates. 6 colored circles are projections of these two sphereplanes. 3 colored dot lines at planes are dividing lines.

$$\overrightarrow{(v_i^t)}^{xy} = (v_i^t)_x^{xy}\,\hat{\imath} + (v_i^t)_y^{xy}\,\hat{\jmath} \tag{11}$$

$$\begin{aligned} \overrightarrow{v_i^t} = {} & ((v_i^t)_x^{xy} + (v_i^t)_x^{xz})\,\hat{\imath} \\ & + ((v_i^t)_y^{xy} + (v_i^t)_y^{yz})\,\hat{\jmath} \\ & + ((v_i^t)_z^{xz} + (v_i^t)_z^{yz})\,\hat{k} \end{aligned} \tag{12}$$

### E. Non-Collision Guarantee

ASCA guarantees that no collision will occur in each time step because the algorithm does not allow agents to enter collision ranges of other agents. In other words, in this algorithm, stopping has more priority than moving in the wrong direction.

The motion acceleration is constant during each time step. The worst-case scenario is when two agents move in opposite directions at their maximum possible speed $v_{max}$. To avoid collision in such scenario, they must be at least $2 v_{max} \Delta t$ apart from each other.

Therefore, if we set the radius of circular avoidance space around each agent to Equation (13) non-collision will be guaranteed.

$$r_i = \frac{d_{min}}{2} + v_{max} \Delta t \tag{13}$$

The real minimum distance that agents must maintain relative to each other is $d_{min}$, but ASCA assumes a minimum distance of $2r_i$ to ensure flawless performance.

### F. Obstacle Avoidance

ASCA is a motion planner algorithm in crowded environments. However, it also has the ability to avoid collision with obstacles. Each obstacle with any geometric shape can be enclosed in a circle for 2D space and can be enclosed in a sphere for 3D space. So, any obstacle can be considered as an agent that has a certain avoidance space diameter. If this additional agent enters the environment while its speed is not adjusted by the algorithm, it is considered as a fixed or moving obstacle with independent speed. All agents assume that this obstacle is an agent and try to avoid that, but the obstacle motion is not managed by ASCA. In a real experiment, this does not matter, because each robot, based on the existing sensors data, runs the algorithm. But in simulation, it is necessary to separate the agents from the obstacles.

### G. Completeness

Under the following conditions, the completeness of the ASCA can be guaranteed. First, all existing obstacles in the environment should be convex. Second, the minimum separation distance between each two pair of final coordinates should be greater than $2d_{min}$. The reason for the first condition is that the proposed method behaves in obstacle avoidance situation similar to the classical path planning BUG algorithms [19], and there is a possibility of dead-lock having non-convex obstacles. The reason for the second condition is that, once the agent reaches its destination and settles there, it will not move from its final position because based on Equation (9), the velocity of the agent is directly related to the distance from final position and in this case the distance is zero. If some agents reach their destinations, they can act as a non-convex obstacle, this may cause a stuck situation for other moving agents that are trying to reach their goal locations. To prevent such situation, there must be at least 2 $d_{min}$ distance between each two final positions.

### H. ASCA pseudocode

ASCA is a distributed algorithm in which all the calculations of each agent are done independently. The initial ($I$) and the final ($F$) position for all agents are the inputs of algorithm and position coordinate ($p_i^t$) and the velocity vector ($v_i^t$) of agent $i$ at each time $t$ for all agents are the outputs. Algorithm 1 illustrates ASCA's pseudocode.

**Algorithm 1    ASCA**

1    $\overline{p_i^0} \leftarrow \overline{I_i}$
2    **for** $t \leftarrow 0$ to $t_{final}$ **step** $\Delta t$ **do** {
3        **for** $i \leftarrow 0$ to $N$ **do in parallel** {
4            $R_i^t = [0, 2\pi]$
5            **for** j$\leftarrow$ 0 to $N$ **do** {
6                **if** $\| \overline{p_i^t} - \overline{p_j^t} \| < d_{min}$ **then** {
7                    $\theta_{ij} = \tan^{-1}(\overline{p_j^t} - \overline{p_i^t})$
8                    $R_i^t = R_i^t \cap [\theta_{ij} + \frac{\pi}{2}, \theta_{ij} + \frac{3\pi}{2}] \}\}$
11        $\overline{d_i^t} = \overline{F_i} - \overline{p_i^t}$
12        $\alpha = \tan^{-1}(\overline{d_i^t})$
13        **if** $\alpha \in R_i^t$ **then** {
14            $\theta_i^t = \alpha$ }
15        **else** {
16            $\theta_i^t = \min(R_i^t)$ }
18        $\overline{v_i^t} = \| \overline{d_i^t} \| \times \binom{\cos(\theta_i^t)}{\sin(\theta_i^t)}$
19        **if** $\| \overline{v_i^t} \| > v_{max}$ **then** {
20            $\overline{v_i^t} = v_{max} \frac{\overline{v_i^t}}{\|v_i^t\|}$ }
22        $\overline{p_i^t} = \overline{v_i^{t-\Delta t}} + \Delta t. v_i^t$
23        $d_{max}^t = \max(d_{max}^t, \| \overline{d_i^t} \|)$ }
25    **if** $d_{max}^t < d_{final}$ **then** {
26        **break** }}

## III.    SIMULATION RESULTS

This section compares ASCA with two state-of-art methods ORCA and FMP. All three methods are real-time, low-cost, distributed, and based on holonomic robots. The video showing all simulations presented in this section is available at https://youtu.be/mkfVihP9DrI.

ASCA implemented in Python without using its parallelization capability but all agents calculate their velocity vectors at each time step in separate loops. The simulator used for visualization purposes also implemented in python using matplotlib library for both 2D and 3D spaces. We used Java implementation of FMP [5], C++ implementation of ORCA [25] and Python implementation of ASCA. We provided the same input for all the algorithms and compared the results. Diversity of implementation languages does not affect the results as the evaluated parameters are independent from the language that is used, because the metric for comparing algorithms in this paper is how long it takes agents move from their initial to final locations (Travel Time) not how long it takes to perform their movement calculations (Run Time). ASCA has not any parameter to set and the FMP and ORCA parameters have not been changed and are set as mentioned in their algorithms.

All tests in this section are performed in a simulation designed using Python and matplotlib. Since all calculations are performed on one system, it is assumed that all agents share their position with each other and the position of agents is calculated using line 22 of Algorithm 1.

We have examined ASCA against two other algorithms in three different tests over the following metrics to evaluate the performance:

**Travel Time:** The total time it takes until the last agent reaches its destination. Calculated as #iterations × $\Delta t$

**Average Trajectory Length:** Average of the trajectory length traversed by all agents.

**Minimum Separation:** The minimum distance between two agents during the algorithm execution. Minimum separation distance is different from the radius of circular avoidance space calculated in Equation (13). This metric should be comparing with $d_{min}$.

In all experiments $N = 100$, $d_{min} = 5m$, $v_{max} = 15 \, m/s$, $\Delta t = 0.02s$, $t_{final} = 200s$ and $d_{final} = 0.05m$ are considered for all three algorithms.

**Table 1**
**Benchmark Test on ASCA, FMP and ORCA**

| Benchmark | Method | Travel Time | Average Trajectory Length | Minimum Separation Distance |
|---|---|---|---|---|
| mirror swap | ASCA | **20.14** | 129 | **5.3** |
| | FMP | - | - | - |
| | ORCA | 28.42 | 141 | 4.9 |
| diagonal swap | ASCA | **34.36** | 231 | **5.0** |
| | FMP | 41.44 | 453 | 4.5 |
| | ORCA | 57.24 | **213** | 4.9 |
| circle swap | ASCA | 86.94 | 1234 | **5.6** |
| | FMP | 40.48 | 464 | 4.7 |
| | ORCA | **37.48** | **382** | 4.6 |
| disk swap | ASCA | **27.62** | 185 | **5.0** |
| | FMP | 29.10 | 316 | 4.7 |
| | ORCA | 30.52 | 253 | 4.8 |
| sphere swap | ASCA | **16.50** | 122 | **5.0** |
| | FMP | 23.86 | 112 | 4.9 |
| | ORCA | 21.92 | **94** | 4.5 |

### A. Benchmark Test

As Table 1 shows, ASCA is compared with FMP and ORCA over five benchmark scenarios. The first three are standard benchmarks for 2D environment introduced in [26]. We add two other benchmarks for dense experiments in 2D and 3D environments.

**Mirror Swap:** A square position set is generated as starting coordinates, and the destination positions are assigned symmetrically with respect to one of the sides of the square.

**Diagonal Swap:** A square position set is generated as starting coordinates, and the destination positions are assigned to each point is symmetrical with respect to a point parallel to the center of the square.

**Circle Swap:** A number of positions are distributed on a circle as starting coordinates, and the destination positions are assigned to each point is symmetric to the ratio of the center of the circle.

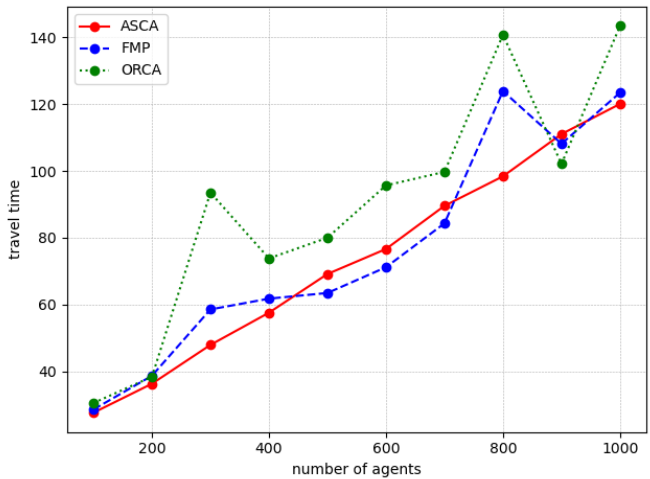

**Fig. 4** Scalability test. Comparison ASCA, FMP and ORCA algorithms travel time over increasing number of agents.

**Disk Swap:** It is like Circle Swap, except that there are some concentric circles with different radiuses.

**Sphere Swap:** A number of positions are distributed on a sphere as starting coordinates, and the destination position assigned to each point is antipodal on the sphere.

As shown in Table 1, while ORCA and FMP violate minimum separation criteria in some cases, the minimum separation parameter for ASCA is not less than $d_{min}$ value in all benchmarks. The reason is that in ASCA, not moving has precedence over making the wrong decision. Another advantage of ASCA over the others is Travel Time. In all benchmarks, except circle swap, the travel time of the proposed algorithm has improved by at least 25% compared to ORCA. ASCA competes with FMP but in most cases it has performed better. Poor performance of ASCA over circle swap can be explained as follows. The decision in ASCA is local, and in this experiment, all the agents are perfectly symmetrical and get closer to each other in the form of a circle, the velocity vector of all agents is selected tangent to the circle at each step. Therefore, the agents tend to rotate in a circle, and since their distance is small, the speed of circle rotation is slow at first, and over the time the rotational speed of the circle reaches to $v_{max}$. Since this speed increasing is slow, the travel time is long.

The average trajectory length represents the energy consumed by the agents and is an important metric in real experiment with robots.

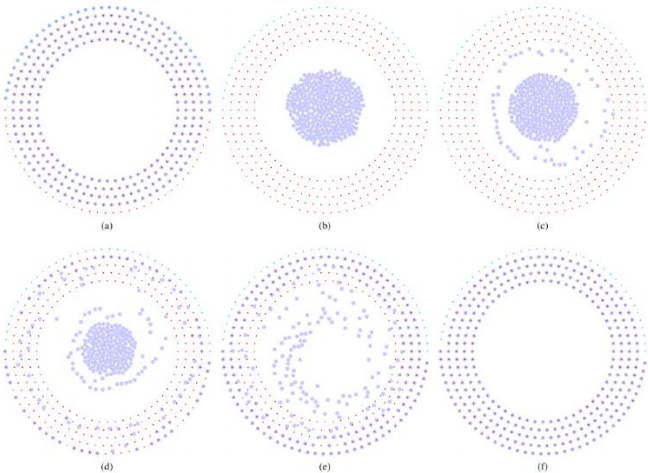

**Fig. 5** ASCA simulation on disk swap for 300 agents. Red dots are final positions and blue circles are the agents (a) t=0s, (b) t=9.58s, (c) t=19.16s, (d) t=28.74s, (e) t=38.32s, (f) t=47.92s.

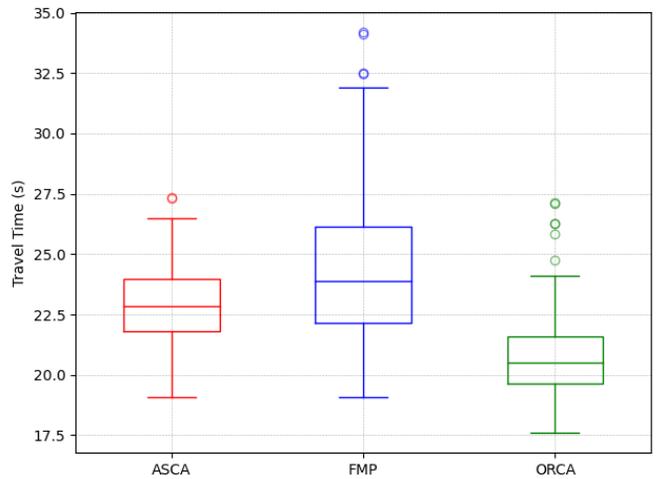

**Fig. 6** Random test. The line inside the box represents the median of 100 travel times. The edges of the box represent the lower and the upper quartiles. The whiskers represent the lowest and the highest within 1.5 times the interquartile range. The small circle points represent the outliers.

Over this evaluation metric, the proposed method is competitive with the others it was compared with. In real experiments, some agents may fluctuate slightly in a confined space, which increases the path length but has a very small effect on the energy consumed (because of limited acceleration) by the agent. Therefore, small differences in average trajectory length metric, can be ignored.

### B. Scalability Test

In this test, we aim to test the ability of ASCA to solve a similar problem with increasing number of agents, and compare the results with two other methods. As represented in Fig. 4, we conducted 10 different experiments on disk swap with the same inner radius and different number of agents that causes different outer radius. As the results show, ASCA in most of the cases outperforms two other methods in terms of travel time. Fig. 5 shows simulation for 300 agents on disk swap benchmark test.

This simulation also reveals that in ASCA, travel time increases linearly by increasing the number of agents. Linear ratio indicates that increasing the size of the problem does not have a detrimental effect on the algorithm. Also, it lets us to predict the travel time for larger number of agents. Based on completeness analysis presents in Section II-G and scalability analysis presented in this section, we can guarantee ASCA eventually converges and its travel time has a linear relationship with the number of agents and the travel time of swarm can be predicted based on the number of agents.

### C. Random Test

In this test, 100 scenarios are generated with 100 pairs of random initial and final points, and all three methods are evaluated over them. As shown in Fig. 6, on average, in 100 random repetitions, the travel time of ORCA is less than ASCA. In all the tests, the starting points are close to the destinations, which has led ORCA to find more optimal paths. But in general, the important issue is the algorithms degree of dependence on the initial and the final positions. For comparison, the aspect ratio metric according to Equation (14) has been used. This metric actually expresses the degree of dependence of the algorithm on the environment. As a result, the position dependence rate is 30% for ASCA, 44% for FMP and 35% for ORCA.

$$Aspect\ Ratio = \frac{\max(travel\ time) - \min(travel\ time)}{\max(travel\ time)} \times 100 \quad (14)$$

## IV. Experimental Results

ASCA is designed to work with Holonomic robots such as quadcopters. For these experiments we used Crazyflie V2.1 [27] which is suitable for running indoor swarm algorithms. Crazyflie does not have the required sensor to detect distance and direction to the neighbors, so it is necessary to use an indoor localization system. We used the HTC Vive BaseStation V1.0 and the lighthouse deck on crazyflies for the localization, which works with infrared light. Although ASCA is a distributed algorithm as the required sensors are not available on crazyflies, so crazyflies need to be able to communicate point-to-point and share their location with their neighbors also p2p communication is not optimally implemented for crazyflies, so we run ASCA on a central computer with i7-3635QM CPU and 8GB RAM, and commands are sent to the 8 crazyflies by 4 crazyradios in real-time.

Fig. 1 shows an experiment performed by 8 crazyflies. In this experiment, $d_{min} = 0.3m$, $v_{max} = 0.5 \ m/s$ and $\Delta t = 0.1s$. No collisions occurred and the execution time was approximately equal to what we expected from the simulation. As shown in this figure, two of the crazyflies that have entered the circle are waiting for the others to move and open their way. Outer 6 crazyflies also move optimally tangentially to the avoidance space. The video of implementation is available at https://youtu.be/mkfVihP9DrI

## V. Conclusion

This paper presents ASCA algorithm to solve the problem of swarm collision avoidance in crowded and complex environments. Our method uses angular calculations to obtain a range of possible motions and ensures that any motions in this interval is collision free.

Simulation results show that ASCA outperforms two other important algorithms in this field, FMP and ORCA, over various parameters such as travel time, average trajectory length and minimum separation. Experimental results using 8 crazyflies confirm the simulation results. We believe that this method can be improved by optimizing and combining next state estimation concepts. As another future work, we plan to provide an ASCA-based approach to work on differential robots. Combining ASCA with other methods to prevent the downwash effect on 3D quadcopter implementations can be considered as another future direction of this research.

## Acknowledgment

The authors would like to thank professor Hugh H.T. Liu from the University of Toronto for his helpful comments on this work.